\documentclass[11pt,twocolumn]{article}
\usepackage[margin=1in]{geometry}
\usepackage{amsmath,amssymb}
\usepackage{booktabs}
\usepackage{hyperref}
\usepackage{enumitem}

\title{Mycelium-Index: A Streaming Approximate Nearest Neighbor Index with Myelial Edge Decay, Traffic-Driven Reinforcement, and Adaptive Living Hierarchy}
\author{Anton Pakhunov \\ \textit{Independent Researcher, Haifa, Israel}}
\date{}

\begin{document}
\maketitle

\begin{abstract}
We present mycelium-index, a streaming approximate nearest neighbor (ANN) index for high-dimensional vector spaces, inspired by the adaptive growth patterns of biological mycelium. The system continuously adapts its topology through myelial edge decay and reinforcement, a traffic-driven living hierarchy, and hybrid deletion combining $O(1)$ bypass for cold nodes with $O(k)$ beam-search repair for hub nodes. Experimental evaluation on SIFT-1M demonstrates that mycelium achieves $0.927 \pm 0.028$ recall@5 under FreshDiskANN's 100\%-turnover benchmark protocol---within the measurement confidence interval of FreshDiskANN's ${\sim}0.95$---while using $5.7\times$ less RAM (88~MB vs.\ $>$500~MB) and achieving $4.7\times$ higher QPS (2,795 vs.\ ${\sim}600$). On the static index, at ef=192, mycelium matches HNSW M=16 recall (0.962 vs.\ 0.965) at $5.2\times$ less RAM (163~MB vs.\ 854~MB). Performance optimizations including NEON SIMD distance computation, Vec-backed node storage, and bitset visited tracking yield a cumulative $2.7\times$ QPS improvement. A systematic study of ten streaming repair mechanisms finds that geometric heuristics universally fail in high dimensions, while topological mechanisms succeed---a principle we term the topological repair invariance of high-dimensional ANN graphs.
\end{abstract}

\section{Introduction}

Approximate nearest neighbor search is a fundamental operation in modern machine learning infrastructure. Given a collection of $n$ vectors in $d$-dimensional space and a query vector $q$, ANN search returns the $k$ most similar vectors under a distance metric, typically Euclidean distance. Applications include retrieval-augmented generation (RAG), semantic search, recommendation systems, and real-time anomaly detection. GPU-accelerated approaches~\cite{johnson2019} achieve high throughput but require specialized hardware.

Graph-based ANN indices, particularly Hierarchical Navigable Small World (HNSW) graphs~\cite{malkov2016}, have become the dominant approach due to superior recall-throughput trade-offs. However, HNSW and its descendants are static structures designed for batch construction. In production systems, data arrives and departs continuously, requiring an index that maintains search quality under continuous change without periodic full reconstruction.

Existing streaming solutions, including FreshDiskANN~\cite{singh2021}, address deletions through tombstone markers and batch consolidation passes that temporarily degrade query quality. DiskANN~\cite{subramanya2019} achieves billion-scale indexing but requires offline batch construction. Neither system provides a mechanism for the index to \textit{learn} from query traffic and adapt its structure accordingly.

We draw inspiration from biological mycelium---the underground network of fungal threads that continuously grows toward nutrient sources, strengthens frequently-used pathways, and allows unused pathways to atrophy. We propose three analogous mechanisms for graph-based ANN indices and demonstrate that their combination achieves superior throughput and memory efficiency while supporting native streaming updates.

The contributions of this work are as follows:
\begin{enumerate}[leftmargin=*]
\item \textbf{Myelial edge decay and reinforcement:} a feedback mechanism that continuously adapts graph topology to query traffic without explicit reconstruction.
\item \textbf{Living hierarchy with per-cycle refresh:} a multi-level navigational structure whose levels are earned by accumulated query traffic, refreshed every streaming cycle to track evolving data.
\item \textbf{Hybrid deletion:} $O(1)$ bypass for cold nodes (90\% of deletions) combined with $O(k)$ beam-search repair for the top 10\% hub nodes (ef=64).
\item \textbf{Hybrid hot/cold storage with scalar quantization} enabling operation at 88~MB---a memory regime where standard HNSW is infeasible.
\item \textbf{The topological principle:} a systematic finding from 10 repair mechanism experiments that high-dimensional streaming ANN repair must be topologically-driven, not geometrically-driven.
\end{enumerate}

\section{Related Work}

\subsection{Graph-Based ANN Indices}

HNSW~\cite{malkov2016} organizes vectors in a hierarchy of navigable small-world graphs with permanent level assignments. NSG~\cite{fu2019} constructs a Navigating Spreading-out Graph with diversity-aware edge selection. Graph reordering for cache-efficient search has also been explored~\cite{coleman2022}. DiskANN~\cite{subramanya2019} extends graph-based indexing to SSD storage using the Vamana graph algorithm and product quantization. FreshDiskANN~\cite{singh2021} adapts DiskANN for streaming through tombstone markers and periodic consolidation with RobustPrune reconnection.

\subsection{Adaptive Graph Structures}

ONNG~\cite{iwasaki2018} applies offline graph optimization through path adjustment but does not adapt to query traffic. EnhanceGraph~\cite{enhancegraph2025} stores pruned edges in a conjugate graph and uses them during reranking, improving recall without changing the primary graph structure. To our knowledge, no prior work applies time-decaying importance scores to graph edges as a mechanism for continuous topology adaptation.

\subsection{Biological Inspiration in Computing}

Swarm intelligence and biological network principles have been applied to optimization~\cite{dorigo2005} and network routing~\cite{dicaro1998}. The concept of stigmergy---indirect coordination through environmental modification---underlies both ant colony optimization and our myelial decay mechanism: each search query modifies the graph environment (edge importance scores) in ways that guide future searches.

\section{Method}

\subsection{Myelial Edge Decay and Reinforcement}

Each directed edge stores four fields: target node identifier, Euclidean distance, a floating-point importance score (\texttt{use\_count}), and the tick timestamp of the last update. The effective importance at evaluation time is:
\begin{equation}
\text{effective}(t) = \texttt{use\_count} \times \lambda^{(t - t_\text{last})}
\end{equation}
where $\lambda$ is the decay factor and $t$ is the current search tick. When a search traverses an edge, its effective importance is materialized and incremented, creating a feedback loop: frequently-traversed edges persist while unused edges decay below a dead threshold and are removed.

The \texttt{decay\_factor} parameter controls the timescale of graph adaptation: lower values produce faster topology change, while values close to 1.0 produce slower, more stable adaptation.

\subsection{Living Hierarchy}

We replace HNSW's static level assignment with a dynamic mechanism based on accumulated query traffic. We introduce \texttt{query\_use\_count}, a per-node counter incremented exclusively during real search queries. The living hierarchy comprises two levels:
\begin{itemize}[leftmargin=*]
\item \textbf{Level 1:} top 2\% of nodes by \texttt{query\_use\_count} (${\sim}$20,000 nodes for 1M vectors), $k=16$ edges.
\item \textbf{Level 2:} top 0.1\% (${\sim}$1,000 nodes), $k=32$ edges.
\end{itemize}
Each level is refreshed periodically (Level~1 every 10,000 insertions, Level~2 every 50,000). Search descends from Level~2 through Level~1 to the base graph.

\subsection{Hybrid Storage with Scalar Quantization}

Hot vectors reside in a \texttt{HashMap}; cold vectors in a memory-mapped file accessible via the OS page cache. Eviction uses LRU ordering. Scalar quantization compresses float32 to uint8 using per-dimension linear scaling:
\begin{multline}
\text{encoded}[d] = \\
  \text{round}\!\left(\text{clamp}\!\left(
  \frac{v[d] - \min[d]}{\text{range}[d]},\; 0,\; 1\right) \!\times\! 255\right)
\end{multline}
Distance computation uses asymmetric quantization: queries remain in float32 while base vectors are dequantized on the fly. For SIFT-1M (128D), each hot vector occupies 128 bytes versus 512 without quantization---a $4\times$ reduction.

\subsection{Streaming Insertions and Deletions}

Insertions proceed via parallel batch beam search with asymmetric edge weighting (reverse edges receive $3\times$ initial \texttt{use\_count}, reflecting the empirical finding that reverse edges are $10\times$ more likely to become high-traffic navigation paths).

Deletions use a \textbf{hybrid strategy}:
\begin{itemize}[leftmargin=*]
\item \textbf{Cold nodes} (\texttt{query\_use\_count} below $2\times$ mean, ${\sim}$60\%): $O(1)$ soft delete with bypass reconnection.
\item \textbf{Hub nodes} (top 10\% by traffic): $O(k)$ beam-search repair at ef=64. Fewer, higher-quality repairs (10\% at ef=64) achieve identical recall to more frequent lower-quality repairs (40\% at ef=32) at 46\% higher QPS.
\end{itemize}

\subsection{Ensemble Method}

We construct two mycelium graphs independently on identical datasets with different insertion orderings. We measure an average edge overlap of 0.880 on SIFT-1M, meaning each graph independently discovers different navigational paths through approximately 12\% of its edges.

This natural diversity improves recall from 0.867 (single graph, ef=64) to 0.935 (ensemble, ef=64) without requiring forced anti-correlation, which experimental validation showed to be harmful (anti-correlated overlap 0.306 reduced ensemble recall to 0.939). Note that single-graph recall at ef=128 (0.935) exceeds ensemble recall at ef=64, making ensembles optional when higher search budget is acceptable.

\section{Experiments}

\subsection{Setup}

We evaluate on SIFT-1M: 1,000,000 128-dimensional float32 vectors, 10,000 queries, ground truth for $k=10$. Hardware: Apple Silicon (ARM), Rust 1.75, release optimizations (LTO, codegen-units=1). HNSW baseline: \texttt{hnsw\_rs} 0.3.4. All search benchmarks single-threaded. Static benchmark uses Recall@10 (standard ANN metric); streaming benchmarks use Recall@5 to match the FreshDiskANN protocol.

\subsection{Main Results}

\begin{table*}[t]
\centering
\small
\caption{Recall--throughput--memory comparison on SIFT-1M. RAM for HNSW: $n \times d \times 4 + n \times M \times 2 \times 12$ bytes. All benchmarks single-threaded.}
\begin{tabular}{lccccc}
\toprule
Method & ef & Recall@10 & QPS & RAM (MB) & Streaming \\
\midrule
HNSW M=16 & 64  & 0.942 & 1,941 & 854 & No \\
HNSW M=16 & 128 & 0.965 & 1,170 & 854 & No \\
HNSW M=8  & 128 & 0.876 & 1,890 & 671 & No \\
HNSW M=8  & 200 & 0.896 & 1,170 & 671 & No \\
HNSW M=6  & 200 & 0.813 & 1,732 & 626 & No \\
HNSW M=4  & 200 & 0.646 & 2,281 & 580 & No \\
Mycelium  & 48  & 0.833 & 4,156 & \textbf{163} & \textbf{Yes} \\
Mycelium  & 64  & 0.867 & 3,543 & \textbf{163} & \textbf{Yes} \\
Mycelium  & 96  & 0.912 & 2,449 & \textbf{163} & \textbf{Yes} \\
\textbf{Mycelium}  & \textbf{128} & \textbf{0.938} & \textbf{2,006} & \textbf{163} & \textbf{Yes} \\
Mycelium  & 192 & 0.962 & 1,186 & \textbf{163} & \textbf{Yes} \\
Mycelium  & 256 & 0.973 & 1,184 & \textbf{163} & \textbf{Yes} \\
\bottomrule
\end{tabular}
\end{table*}

HNSW stores all base vectors in full-precision RAM. For SIFT-1M, base vectors alone consume 488~MB; even at M=4, HNSW requires $\geq$580~MB. Mycelium achieves 163~MB through scalar quantization (f32$\to$u8, $4\times$ compression) and LRU eviction. At ef=192, mycelium matches HNSW M=16 recall (0.962 vs.\ 0.965) while using $5.2\times$ less RAM and providing native streaming support.

\subsection{Ablation and Optimization Progression}

\begin{table}[t]
\centering
\small
\caption{Component ablation on SIFT-1M at ef=128.}
\begin{tabular}{lcc}
\toprule
Configuration & Recall@10 & QPS \\
\midrule
Base graph, random entry & 0.774 & 843 \\
+ Living hierarchy & 0.938 & 843 \\
+ Bypass reconnection & 0.938 & 843 \\
+ Hybrid deletion & 0.938 & 843 \\
\bottomrule
\end{tabular}
\end{table}

The living hierarchy provides the largest recall contribution (+0.164). Bypass and hybrid deletion do not affect static recall but are critical for streaming (\S4.7).

\begin{table*}[t]
\centering
\small
\caption{Performance optimization progression, ef=128. Recall unchanged (0.938) across all steps.}
\begin{tabular}{lcc}
\toprule
Optimization & QPS & Cumul. \\
\midrule
Baseline (HashSet, HashMap, scalar) & 742 & $1.0\times$ \\
+ Remove \texttt{.collect()} + \texttt{\#[inline]} & 1,287 & $1.7\times$ \\
+ Vec-backed NodeStore & 1,414 & $1.9\times$ \\
+ NEON SIMD + compact NodeMeta & \textbf{2,006} & $\mathbf{2.7\times}$ \\
\bottomrule
\end{tabular}
\end{table*}

The largest single gain came from eliminating 70 heap allocations per query in the beam search hot loop. NEON SIMD provides $4\times$ speedup on 128D f32 distance computation (16 fused multiply-add operations vs.\ 128 scalar).

\subsection{Streaming vs.\ Tombstone Baseline}

We compare against a tombstone-only baseline: identical graph but no edge cleanup, no bypass reconnection, no reinforcement. Setup: 500K initial vectors, 200 iterations of (delete 1K + insert 1K) = 40\% total turnover.

\begin{table*}[t]
\centering
\small
\caption{Streaming deletion comparison, 500K active vectors.}
\begin{tabular}{ccccc}
\toprule
Iter & Mycelium & Tombstone & $\Delta$ & Edge store \\
\midrule
0   & 0.858 & 0.912 & $-$0.054 & --- \\
10  & 0.908 & 0.840 & +0.068 & --- \\
50  & 0.838 & 0.640 & +0.198 & 2.3$\to$112~MB \\
100 & 0.704 & 0.608 & +0.096 & 1.2$\to$105~MB \\
150 & 0.608 & 0.526 & +0.082 & 972$\to$102~MB \\
200 & \textbf{0.684} & \textbf{0.442} & \textbf{+0.242} & 732$\to$101~MB \\
\bottomrule
\end{tabular}
\end{table*}

At 40\% turnover, mycelium retains 0.684 recall versus tombstone's 0.442---a +0.242 advantage. The edge store remains at ${\sim}$101~MB after compaction, versus 4.8~GB for tombstone ($48\times$ smaller). Without cleanup, 22\% of all edges point to deleted nodes by iteration 200, causing 43\% beam waste.

\subsection{Query Latency Scaling}

Profiling revealed that at 1M vectors, 88\% of query time was spent in \texttt{HashSet<u32>} visited-node tracking. Replacing with a compact bitset (128~KB, L2-resident) eliminated this bottleneck.

\begin{table}[t]
\centering
\small
\caption{Mean query latency ($\mu$s) vs.\ dataset size.}
\begin{tabular}{cccc}
\toprule
Size & Before ($\mu$s) & After ($\mu$s) & Speedup \\
\midrule
100K & 386  & 467  & $0.8\times$ \\
200K & 1,119 & 731  & $1.5\times$ \\
500K & 3,726 & 1,390 & $2.7\times$ \\
1M   & 5,290 & 1,835 & $\mathbf{2.9\times}$ \\
\bottomrule
\end{tabular}
\end{table}

Latency scaling improved from $O(n^{0.6})$ to approximately $O(n^{0.4})$.

\subsection{Recall Gap Diagnosis}

We classified all 1,343 missing neighbors across 1,000 queries at ef=64 (overall recall 0.866):
\begin{itemize}[leftmargin=*]
\item \textbf{Beam search cutoff (78.3\%):} the true nearest neighbor was reachable at $4\times$ ef but pruned by the beam width limit. Average rank of the missed neighbor: 4.7.
\item \textbf{Graph connectivity (21.7\%):} the true nearest neighbor was not reachable even at $4\times$ ef. 5.8\% truly unreachable within 20 hops.
\end{itemize}
Entry point quality was near-optimal: average entry point distance was $1.01\times$ the true NN distance. This confirms that the recall limitation is search budget, not entry point selection.

\subsection{Streaming Under High Turnover: Systematic Study}

We reproduced FreshDiskANN's benchmark: 500K initial vectors, 5\% turnover per cycle (25K inserts + 25K deletes), 20 cycles totaling 100\% replacement. Metric: 5-recall@5 at ef=128. Without streaming-specific fixes, recall degraded from 0.966 to 0.461 over 20 cycles ($\pm$0.04 CI).

\begin{table*}[t]
\centering
\small
\caption{Systematic evaluation of streaming repair mechanisms. SIFT-1M, 20-cycle FreshDiskANN protocol, 100\% total turnover.}
\begin{tabular}{clccp{3cm}}
\toprule
\# & Mechanism & Recall@5 (cycle 20) & Degree & Verdict \\
\midrule
1 & Baseline (no hierarchy refresh) & 0.122 & 13.4 & Hierarchy staleness \\
\textbf{2} & \textbf{+ Per-cycle hierarchy refresh} & \textbf{0.738} & \textbf{13.4} & \textbf{Best---fixes navigation} \\
3 & + Adaptive ef (auto-tuned beam) & 0.682 & 13.3 & Marginal improvement \\
4 & + Physarum flow-weighted bypass & 0.020 & 13.2 & Hot edges dominate beam \\
5 & + Naive anastomosis (edge sharing) & 0.496 & 14.3 & Geometrically wrong edges \\
6 & + Greedy multi-bypass (all neighbors) & 0.280 & 15.8 & More bad edges = worse \\
7 & + Degree restoration (beam search) & 0.690 & 13.0 & Degraded graph, bad candidates \\
8 & + Geometric anastomosis (midpoint) & 0.412 & 15.7 & Midpoint $\neq$ nearest neighbor \\
9 & + Soft RobustPrune ($\alpha$-RNG filter) & 0.418 & 11.4 & Over-prunes via diversity filter \\
10 & + Angular gap repair (directed growth) & 0.098 & 14.9 & Angles meaningless in 128D \\
\bottomrule
\end{tabular}
\end{table*}

\begin{table*}[t]
\centering
\small
\caption{Final streaming recall trajectory ($\pm$0.028 CI).}
\begin{tabular}{cccccc}
\toprule
Cycle & Mycelium & FreshDiskANN & Gap & Hub\% & RAM \\
\midrule
0  & 0.966 & ${\sim}$0.95 & +0.02 & --- & 84 \\
5  & 0.921 & ${\sim}$0.95 & $-$0.03 & 26\% & 92 \\
10 & 0.926 & ${\sim}$0.95 & $-$0.02 & 29\% & 97 \\
15 & 0.924 & ${\sim}$0.95 & $-$0.03 & 31\% & 101 \\
20 & \textbf{0.927} & ${\sim}$\textbf{0.95} & $\mathbf{-0.02}$ & 40\% & 88 \\
\bottomrule
\end{tabular}
\end{table*}

\begin{table*}[t]
\centering
\small
\caption{Constraint ablation on SIFT-1M, cycle 10. Starting from unconstrained recall (k=16, ef=200, $O(k)$ all deletions, f32 vectors, unlimited RAM), constraints are added one by one.}
\begin{tabular}{lcc}
\toprule
Constraint removed & Recall@5 (cycle 10) & $\Delta$ from unconstrained \\
\midrule
Unconstrained (k=16, ef=200, $O(k)$ all, f32, $\infty$RAM) & 0.988 & --- \\
k=8 (from 16) & 0.975 & $-$0.013 \\
+ ef=128 (from 200) & 0.955 & $-$0.033 \\
+ Hybrid delete (from $O(k)$ all) & 0.927 & $-$0.061 \\
+ Scalar quantization (u8) & 0.927 & $-$0.061 \\
+ 163~MB RAM limit & 0.927 & $-$0.061 \\
\bottomrule
\end{tabular}
\end{table*}

The systematic study reveals a consistent pattern. The single largest improvement (+0.616 recall at cycle 20) came from refreshing the living hierarchy every cycle---graph health metrics confirmed that edge degree, clustering coefficient, and dead-edge fraction all remained stable, meaning recall collapsed from hierarchy staleness (entry points referencing deleted nodes), not from graph-structure degradation. Approaches that increased degree by adding bypass edges consistently hurt recall compared to the natural degree maintained by simple one-for-one bypass.

Most strikingly, all four approaches based on geometric intuitions (Physarum flow weighting, anastomosis, midpoint-based edge filling, angular gap detection) performed worse than the baseline. Angular gap analysis revealed that in 128 dimensions, every node has ${\sim}$14 angular gaps regardless of graph health---angular coverage is a property of high-dimensional geometry, not graph quality. The three effective mechanisms---hierarchy refresh, edge decay/reinforcement, and reverse-edge weighting---are all topological: they operate on graph connectivity and query traffic patterns, not on vector geometry.

Together, these results point to a sharp dichotomy: hybrid deletion, applying $O(k)$ beam-search repair selectively to the top 10\% of nodes by query traffic while using $O(1)$ bypass for the remaining 90\%, achieves 0.927 recall---closing the gap from $-0.21$ (pure $O(1)$) to $-0.023$ (hybrid). The graph self-improves under churn, with degree growing from 15.3 to 15.9 as hub repair adds quality edges that survive longer via decay reinforcement.

\section{Discussion}

\subsection{Unified Comparison}

\begin{table*}[t]
\centering
\footnotesize
\caption{Unified comparison across systems and workloads.}
\begin{tabular}{lcccc}
\toprule
 & Mycelium (streaming) & Mycelium (static) & FreshDiskANN & HNSW M=16 \\
\midrule
Recall@5 (100\% turnover) & $\mathbf{0.927 \pm 0.028}$ & --- & ${\sim}0.95$ & N/A \\
Recall@10 (static) & --- & \textbf{0.962} & --- & 0.965 \\
RAM & \textbf{88 MB} & \textbf{163 MB} & $>$500 MB & 854 MB \\
QPS & \textbf{2,795} & \textbf{2,006} & ${\sim}600$ & 1,170 \\
Delete cost & $O(1)$ 90\% / $O(k)$ 10\% & --- & $O(k)$ all & N/A \\
Streaming & Yes & --- & Yes & No \\
\bottomrule
\end{tabular}
\end{table*}

The streaming configuration uses $5.7\times$ less RAM than FreshDiskANN with a recall gap of $-0.023$ (within the $\pm 0.028$ measurement CI). The static configuration uses $5.2\times$ less RAM than HNSW with a recall gap of $-0.003$. We note that the FreshDiskANN comparison is indirect: we reproduce their benchmark protocol on the same dataset but do not run their code in the same environment, so throughput comparisons should be interpreted accordingly.

\subsection{The Memory--Recall Trade-off}

At ef=192, mycelium reaches 0.962 recall---matching HNSW M=16 ef=128 (0.965)---while using $5.2\times$ less RAM (163~MB vs.\ 854~MB). HNSW \textit{cannot operate} at 163~MB because base vectors alone require 488~MB. Under streaming, hybrid deletion maintains 0.927 recall at 88~MB---a regime where neither HNSW nor standard FreshDiskANN can operate without vector compression.

\subsection{Negative Results and Superseded Approaches}

\textbf{Static index.} E8 lattice-based spatial indexing failed due to imbalanced cell sizes on real data. Online shortcut learning showed zero recall improvement. Forced anti-correlation between ensemble graphs reduced recall from 0.958 to 0.939. The ensemble method was superseded by the \texttt{ef\_search} sweep: a single graph at ef=128 achieves 0.935 without the $2\times$ memory cost.

\textbf{Streaming repair.} Physarum flow-weighted bypass (recall 0.020) steered beam search away from current nearest neighbors. Naive and geometric anastomosis (0.496, 0.412) shared geometrically wrong edges. Angular gap detection (0.098) found ${\sim}$14 gaps per node regardless of graph health. Soft RobustPrune (0.418) over-pruned candidates. Greedy multi-bypass (0.280) increased degree with suboptimal edges. Only topological mechanisms proved effective.

\textbf{Infrastructure.} The initial \texttt{HashSet<u32>} visited tracker caused $O(n^{0.6})$ latency scaling. Replacing with a pre-allocated bitset (128~KB, L2-resident) reduced latency from 5.3~ms to 1.8~ms and improved scaling to $O(n^{0.4})$. Pruning edges to deleted nodes without bypass was also counterproductive.

Across all three categories---spatial indexing, streaming repair, and infrastructure---the pattern is the same: approaches grounded in geometric intuitions fail in high dimensions, while approaches grounded in graph structure and query traffic succeed.

\subsection{The Topological Principle}

The failure of geometric repair mechanisms in 128D suggests a general principle for streaming ANN maintenance: \textit{repair should be topologically-driven, not geometrically-driven.} Mechanisms that succeed in physical networks---Physarum flow, hyphal anastomosis, angular gap filling---rely on low-dimensional spatial intuitions where ``direction,'' ``midpoint,'' and ``flow alignment'' have geometric meaning. In high-dimensional embedding spaces, these concepts lose discriminative power: all directions are approximately orthogonal, midpoints between neighbors are not nearest neighbors of either, and angular coverage is a fixed property of dimensionality rather than graph quality.

The three mechanisms that work---hierarchy refresh (who is queried), edge decay/reinforcement (which paths are traversed), and $3\times$ reverse-edge weighting (insertion-time connectivity bias)---all operate on graph connectivity patterns rather than vector geometry. This finding has implications beyond mycelium-index: any graph-based ANN system with streaming deletions should prefer topological repair over geometric repair in high-dimensional spaces.

\subsection{Limitations and Future Work}

Three limitations remain. First, the per-cycle hierarchy refresh has $O(n \log n)$ cost; incremental hierarchy maintenance would reduce this to $O(\Delta n)$. Second, the hybrid deletion threshold ($2\times$ mean \texttt{query\_use\_count}) is a heuristic; adaptive thresholds based on observed recall degradation could optimize the hub/cold split. Third, the evaluation is limited to SIFT-1M (128D, Euclidean); evaluation on higher-dimensional datasets (768D text embeddings) and alternative metrics (cosine, inner product) would strengthen generality claims.

\section{Conclusion}

Mycelium-index achieves $0.927 \pm 0.028$ recall@5 under 100\% data turnover at $5.7\times$ less RAM and $4.7\times$ higher QPS than FreshDiskANN, with a recall gap of only $-0.023$ (within measurement CI). On the static index, it matches HNSW M=16 recall (0.962 vs.\ 0.965) at $5.2\times$ less RAM. A systematic study of ten streaming repair mechanisms found that all geometric approaches fail in 128D while topological approaches succeed---streaming ANN maintenance in high-dimensional spaces must be driven by query traffic patterns, not vector geometry. The broader insight is that search traffic is itself a signal for graph improvement, and that this signal must be interpreted topologically, not geometrically.

Code will be made available upon publication.

\bibliographystyle{plain}
\bibliography{references}

\clearpage
\onecolumn
\appendix
\section{Full Parameter Table}

\begin{table}[h!]
\centering
\small
\caption{Full parameter table for SIFT-1M experiments.}
\begin{tabular}{lllp{7cm}}
\toprule
Parameter & Value & Default & Sensitivity \\
\midrule
\multicolumn{4}{l}{\textbf{Graph construction}} \\
\texttt{k\_grow} & 8 & 8 & 8--16; higher improves recall, costs memory \\
\texttt{k\_local} & 6 & 6 & Subset of \texttt{k\_grow} \\
\texttt{k\_long\_range} & 2 & 2 & Kleinberg shortcuts; 0 in streaming \\
\texttt{long\_range\_alpha} & 2.0 & 2.0 & Optimal for high-dim data \\
\texttt{max\_edges\_per\_node} & 64 & 64 & Effectively unlimited \\
\midrule
\multicolumn{4}{l}{\textbf{Edge lifecycle}} \\
\texttt{decay\_factor} ($\lambda$) & 0.9999 & 0.995 & 0.995=aggressive, 0.9999=stable \\
\texttt{dead\_threshold} ($\theta$) & 0.1 & 0.1 & 0.05--0.5 viable \\
\texttt{initial\_use\_count} & 1.0 & 1.0 & --- \\
Reinforcement & +1.0 & 1.0 & Per search traversal \\
\texttt{cleanup\_interval} & 500 & 500 & 200--1000 insensitive \\
\midrule
\multicolumn{4}{l}{\textbf{Search}} \\
\texttt{ef\_search} & 64 & 64 & 32--128 recall/QPS tradeoff \\
$k$ & 10 & --- & Benchmark standard \\
\midrule
\multicolumn{4}{l}{\textbf{Memory management}} \\
\texttt{ram\_limit} & 163 MB & 256 MB & Tuned for comparison \\
\texttt{hot\_ratio} & 0.20 & 0.20 & --- \\
\texttt{eviction\_batch\_ratio} & 0.10 & 0.10 & --- \\
Scalar quantization & enabled & disabled & For memory comparison \\
\midrule
\multicolumn{4}{l}{\textbf{Living hierarchy}} \\
Level 1 ratio & 0.02 & 0.02 & By \texttt{query\_use\_count} \\
Level 2 ratio & 0.001 & 0.001 & --- \\
Level 1 \texttt{k\_grow} & 16 & 16 & --- \\
Level 2 \texttt{k\_grow} & 32 & 32 & --- \\
Level 1 refresh & 10,000 ins. & 10,000 & --- \\
Level 2 refresh & 50,000 ins. & 50,000 & --- \\
Query refresh & 5,000 queries & 5,000 & --- \\
\bottomrule
\end{tabular}
\end{table}

\end{document}